\documentclass{IOS-Book-Article}

\usepackage{mathptmx}
\usepackage{soul}\setuldepth{article}
\usepackage{graphicx}
\usepackage{graphics} 
\usepackage{subcaption}
\usepackage{mathptmx}

\def\hb{\hbox to 11.5 cm{}}

\begin{document}

\pagestyle{headings}
\def\thepage{}

\begin{frontmatter}

\title{A Multi-Sensor Dataset for Monitoring the Operational Environment of\\ Rail Vehicles}

\markboth{}{September 2025\hb}

\author[A]{Claudio Diotallevi}%
 \author[A]{ Rodrigo Gudiño}
 \author[A]{Zaharia Pachalieva}
 \author[B]{Philipp Neumaier\thanks{Corresponding Author: Philipp Neumaier, E-mail: philipp.neumaier@deutschebahn.com.}}
 \author[B]{Patrick Naumann}
 \author[B]{Erik Bochinski}
 \author[B]{Volker Eiselein}
and
\author[B]{Martin Köppel}

\runningauthor{B.P. Manager et al.}
\address[A]{understandAI GmbH, An der RaumFabrik 33a, 76227 Karlsruhe, Germany}
\address[B]{DB InfraGO AG, EUREF-Campus 17, 10829 Berlin, Germany}

\begin{abstract}
Reliable environment monitoring is essential for the safe and efficient operation of automated railway systems, covering all Grades of Automation (GoA), from partially automated (GoA2) to fully automated operation (GoA4). Artificial Intelligence (AI) plays a central role in enabling these systems to detect, classify, and react to potential hazards in real time. The development of such AI-based perception systems requires large volumes of accurately annotated data for training and validation. 

Within the Digitale Schiene Deutschland (DSD) program,
DB InfraGO AG and understandAI GmbH have developed a comprehensive multi-
sensor dataset tailored to the needs of railway environment perception. 
This dataset contains over 7 million high-quality annotations of both railway-specific and general perception objects, captured under varying operational scenarios.  
The finalized dataset can now be requested at the DB InfraGO AG and serve as a valuable resource for advancing AI-driven environment monitoring in the railway domain.
\end{abstract}

\end{frontmatter}
\markboth{September 2025\hb}{September 2025\hb}

\section{Introduction}
The monitoring of the route and track environment plays an important role in automated driving in the railway industry. For example, this technology can be used as an assistance system for route monitoring in automation level GoA2, where the train driver is still on board. In fully automated, driverless driving at automation level GoA4, these systems finally take over environment monitoring completely independently. With the help of artificial intelligence (AI), they react automatically to risks and dangerous events on the route. 
Nevertheless, very large amounts of data are required to develop such AI processes for environment perception. So-called annotations provide additional information that describes the relevant objects in the data. 
In 2022, Digitale Schiene Deutschland (DSD) \cite{r1} has set up the "Data Factory" \cite{r1_0} a comprehensive data ecosystem that ensures the flow of data from the vehicle to the trackside and into the backend and the cloud. It encompasses data collection, processing, quality assurance, ML/AI training, storage, management, analysis, and provision. 

As part of the DSD sector initiative, DB InfraGO AG has furthermore worked together with understandAI GmbH (UAI) \cite{r2}, a company specializing in this field, to create a comprehensive multi-sensor dataset. The dataset contains over 7 million annotations of railway-specific and generic perception objects. The work on the dataset has been completed and can now be made available to the industry for the development of environment monitoring systems\footnote[1]{The dataset is available upon request from DB InfraGO AG by contacting either \mbox{philipp.neumaier@deutschebahn.com} or \mbox{martin.koeppel@deutschebahn.com}.}.

\section{Existing datasets for computer vision in railways}
\label{rvd}
\begin{table}
  \centering
  \caption{Existing public datasets for CV in railways.}
  \begin{tabular}{|l||r|r|r|}
    \hline
    \textbf{Dataset}& \textbf{Year} & \textbf{Frames} & \textbf{Sensors} \\
    \hline
     RailSem19 \cite{r3}& 2019 & $8500$   & camera \\
     FRSign \cite{r4}      & 2020 & $105352$ & camera \\
     RAWPED \cite{r5}      & 2020 & $26000$  & camera \\  
     Rail-DB \cite{r6}     & 2022 & $7432$   & camera \\
     RailSet \cite{r7}    & 2022 & $6600$   & camera \\
     GERALD \cite{r8}      & 2023 & $5000$   & camera \\
     OSDaR23 \cite{r9}      & 2023 & $1534$   & camera, lidar, radar \\   
     RailGoerl24 \cite{r10}   & 2024  & $12205$ & camera, lidar \\
     SynDRA \cite{r11}   & 2025  & $6572$ & camera (simulated) \\
    \hline
  \end{tabular}
  \label{exdrv}
\end{table}
In Table \ref{exdrv}, publicly available computer vision datasets are listed which captured the environment of railway trains with frontal onboard sensors. These datasets consist of annotated sensor frames extracted from sensors such as camera, lidar and radar.

\mbox{RailSem19} features scenes of railways and trams from 38 countries, providing annotations as geometric shapes and dense, pixel-level semantic segmentations for elements such as trains, switches, switch states, platforms, buffer stops, rail traffic signs, and railway signals. 
RailSet offers similar annotations to RailSem19 but also includes railway scenes with both normal and anomalous conditions, such as rail discontinuities and hole anomalies. Rail-DB contains annotated recordings of high-speed rail infrastructure.
\mbox{FRSign} is a dataset with frames annotated with bounding boxes of French railway signals in different states, while GERALD includes frames featuring German railway signals. RAWPED consists of frames annotated with bounding boxes for pedestrians.
\mbox{OSDaR23} presents a multi-sensor dataset of 45 sequences that were created to foster driverless train operation on mainline railways. The sensor setup consists of multiple calibrated and synchronized infrared (IR) and visual (RGB) cameras, lidars, a radar, and position and acceleration sensors mounted on the front of a rail vehicle. The dataset represents the first publicly available multi-sensor dataset for the railway environment.
\mbox{RailGoerl24} presents a camera based dataset recorded in a railway test center of TÜV SÜD Rail, in Görlitz, Germany. 
\mbox{RailGoerl24} also includes a terrestrial LiDAR scan covering
parts of the area used to acquire the RGB camera data.
\mbox{SynDRA} provides synthetic photo-realistic images with precise
pixel-level annotations. The design of these environments
was carefully planned. Furthermore, the
generation pipeline allows an easy extensions for new scenarios and the integration of other annotations.

In addition to these open datasets, there are others that have been used in research but are not publicly available, such as the RAILO dataset, which comprises 4,651 manually annotated single RGB-camera frames from six different scenes \cite{r11}. A further non-public dataset was used in \cite{r13} and \cite{r14} to evaluate active learning and track detection methods. 

\section{Rail vehicles for recording multi-sensor data}
In order to record suitable data, two rail vehicles were equipped with sensors.

\subsection{Track maintenance vehicle}

\begin{figure}[t]
      \includegraphics[width=\linewidth]{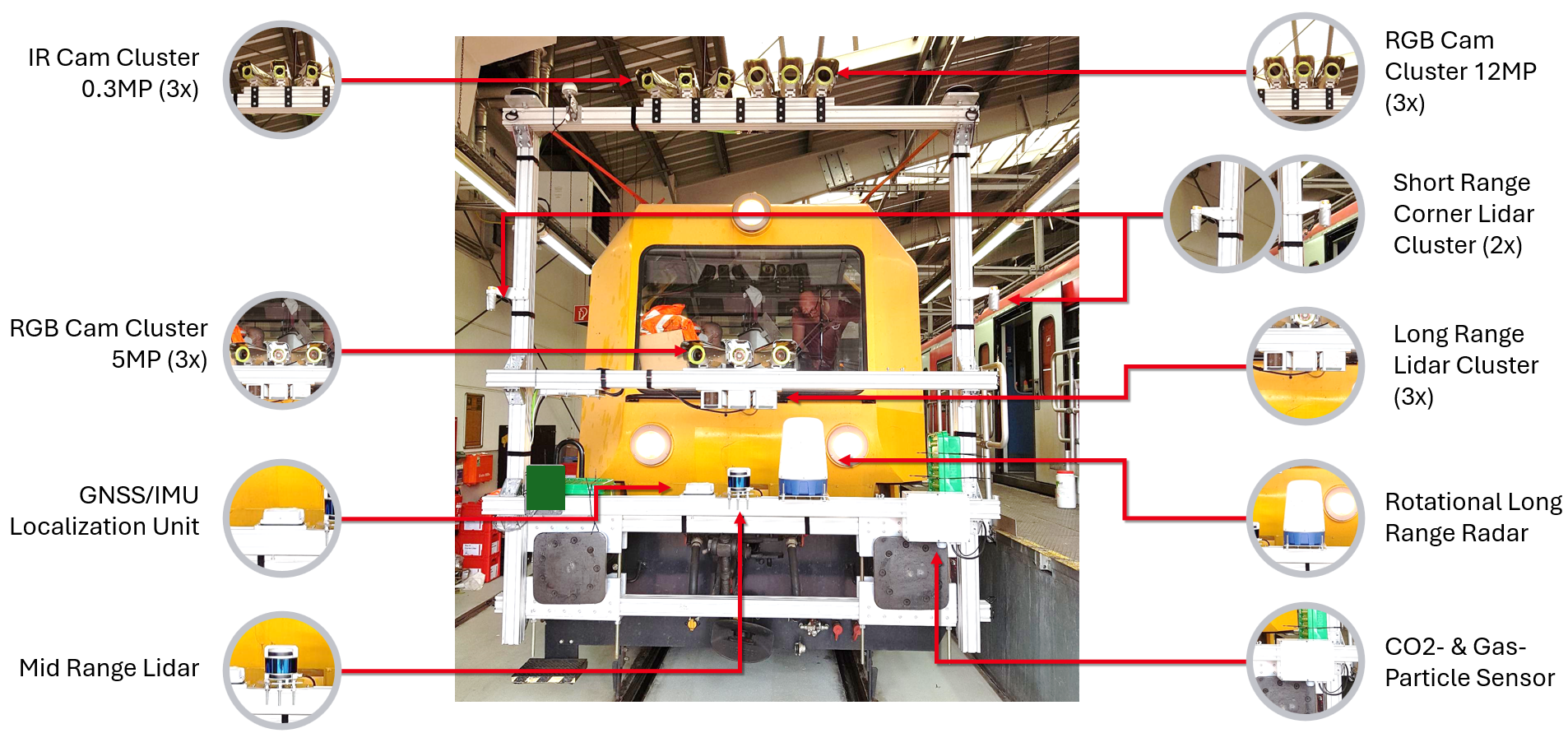}
      \caption{Track maintenance vehicle (GAF).}
      \label{abb-gaf}
\end{figure}

\begin{figure}[t]
      \includegraphics[width=0.6\textwidth]{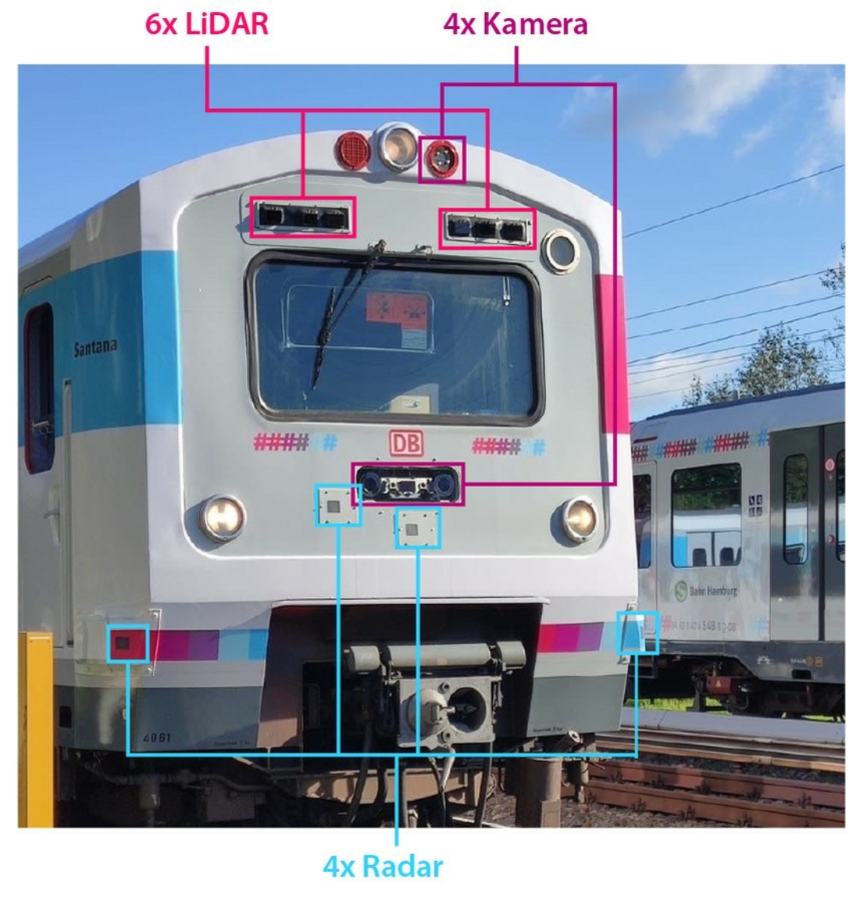}
      \caption{Commuter train (BR472).}
      \label{abb-s4r}
\end{figure}

The first test platform was a track maintenance vehicle of the type GAF, as illustrated in Fig. \ref{abb-gaf}. As part of an internal research and development initiative led by DB InfraGO AG, this vehicle was equipped with a variety of sensors mounted on its front section. The goal of the project was to explore the suitability of different sensor modalities for environment perception in railway monitoring.

The annotated sensor configuration includes a total of six RGB (color) cameras and three infrared (IR) cameras for visual data acquisition. Additionally, six LiDAR sensors with varying ranges and detection capabilities were installed to capture high-resolution spatial information. A radar sensor was also integrated to enhance object detection performance under challenging weather or lighting conditions. The setup also includes position and acceleration sensors to provide accurate localization and motion context and a gas-particle sensor. The vehicle is depicted in Fig. \ref{abb-gaf} and further outlined in \cite{r9}.

The sensor data was collected during multiple test runs conducted in Hamburg and Berlin, Germany. These recordings span a wide range of operational conditions, including both regular and non-regular scenarios, enabling the development and evaluation of robust perception algorithms.

\subsection{Commuter vehicle (Hamburg)}

The second test vehicle (Fig. \ref{abb-s4r}) was equipped with a comprehensive sensor setup during the Sensors4Rail research and development project \cite{r21}. This project brought together industry partners including Deutsche Bahn, Bosch Engineering, HERE Technologies, Ibeo Automotive Systems (now MicroVision), and Siemens Mobility, with the shared objective of advancing environment perception technologies for future rail automation.

Within the scope of this project, a BR472 commuter train was selected as the test platform. The vehicle was equipped with a sensor configuration designed to capture a variety of environmental and operational data. The setup included three RGB (color) cameras for visual scene understanding, one infrared (IR) camera to support perception under low-light or poor visibility conditions, six LiDAR sensors with varying fields of view and detection ranges for detailed 3D spatial mapping, and four radar units to enhance object detection and tracking, especially under adverse weather conditions.

The data acquisition took place along a 23-kilometer section of the S21 commuter rail line, specifically between Hamburg-Berliner Tor and Bergedorf (Hamburg, Germany). This urban-to-suburban route provided a wide range of operating conditions and infrastructure elements. It was suitable for testing the performance and reliability of the systems under realistic railway environments.

\section{Data annotations and object classes}

\begin{figure}[t]
      \includegraphics[width=\textwidth]{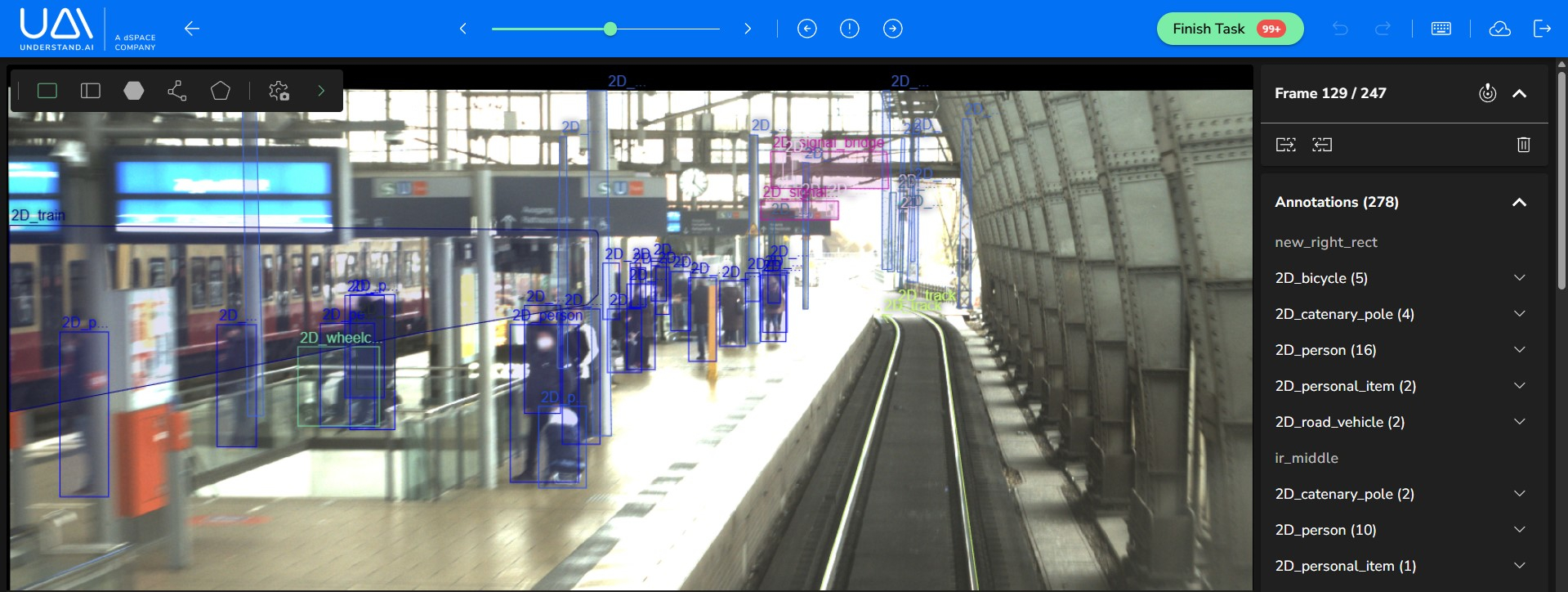}
      \caption{2D-Bounding Boxes of persons in the annotated camera image.}
      \label{abb-1}
\end{figure}

\begin{figure}[t]
      \includegraphics[width=\textwidth]{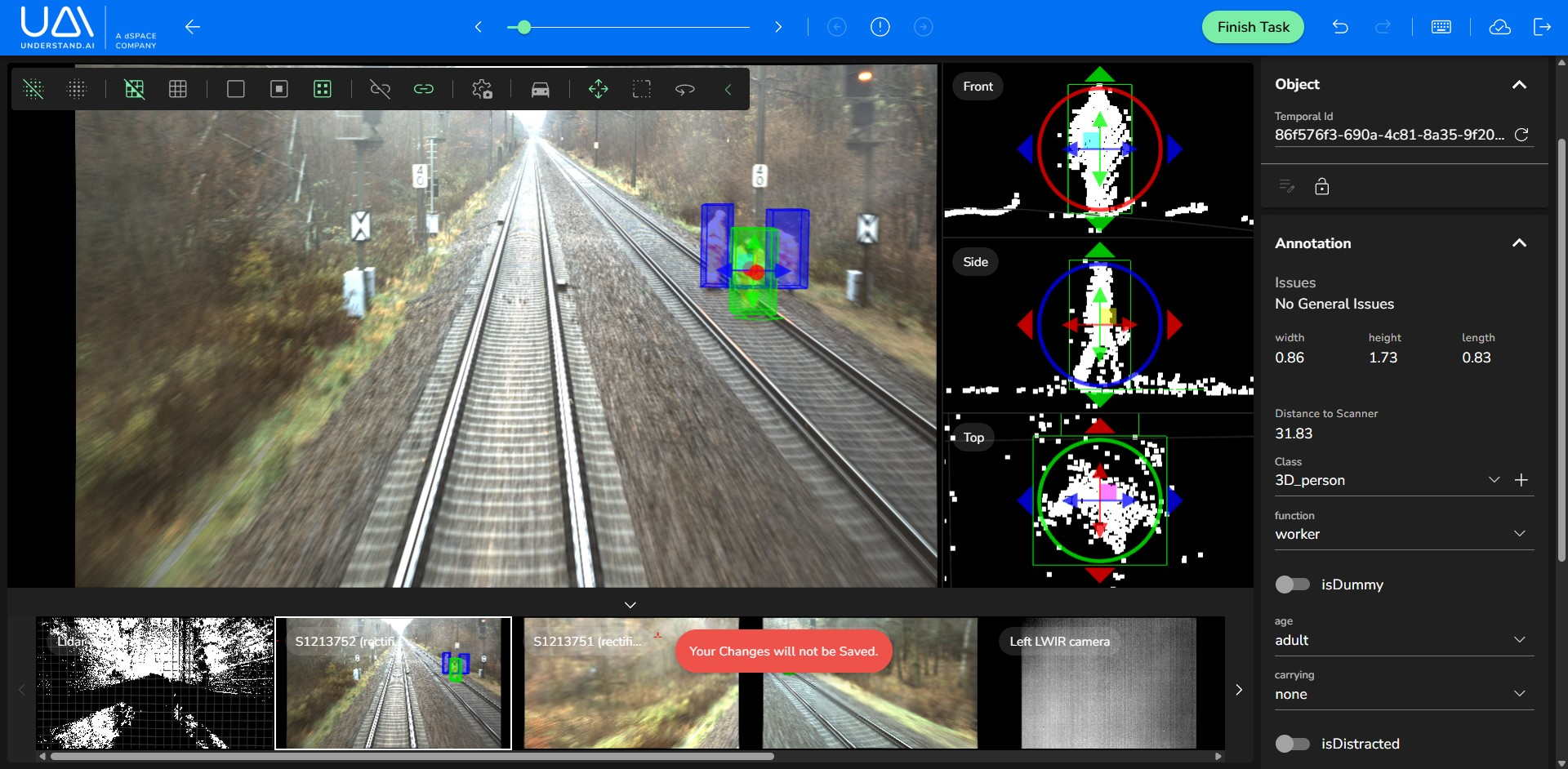}
      \caption{3D-bounding boxes of workers, projected into the camera image.}
      \label{abb-2}
\end{figure}

Data annotation involves the marking of object classes in the sensor data (Fig. \ref{abb-1}, \ref{abb-2}, \ref{annotation-train} and \ref{annotation-train-3D}). There are different types of annotations, e.g. 2D or 3D bounding boxes, segmentations, polygons and polylines. The annotations also contain additional information about the objects via attributes. Since 2021, UAI and DB InfraGO AG have jointly created annotations for 21 object classes.
The utilized annotation methodology included both automatic software tools and manual annotation. The software tools were used in all project phases (data import, data quality checks, annotation and data export) to increase efficiency and thus reduce manual annotation times. 

All data was first annotated by experts in the 3D point clouds recorded by the lidar (Fig. \ref{abb-2}). These annotations were then projected onto the other sensors using projection functions. The various projection functions are outlined below:

\begin{enumerate}
    \item Projection of 3D-bounding boxes to 2D-bounding boxes.
    \item Projection of 3D-bounding boxes to 2D-Polygons.
    \item Projection of 2D-RGB cameras to 2D-Infrared (IR).
    \item Projection of 3D-point clouds to radar images.
\end{enumerate}

All projected annotations were then checked and reworked if necessary. During the annotation projects, UAI was able to realize a peak delivery throughput of 140,000 annotations/week.

\section{Quality control}
Strict data quality management was crucial to achieve the quality objectives. The project team had utilized multiple levels of quality control, from the initial quality assessment of the raw data, through prototype development, automated quality assurance in production, to final quality assurance before or after delivery. The various steps are described below:

\begin{enumerate}
    \item Quality assessment of raw data: Tools and manual checks were applied to detect early quality issues of the raw data (calibration, odometry, etc.) that could potentially lead to delays and costly re-work.
    \item Prototype development: For each new version of the annotation specifications or sensor configuration, the team performed several iterative loops to verify compliance with the specification using sample datasets. Only then did they start the annotation process on a production scale.
    \item Automated QA validators for data annotations: Automated software validators were applied as real-time support to reduce errors during annotation to minimize the need for extensive checks and shorten the time to completion.
    \item Pre-delivery quality assurance: Final checks were carried out before data delivery to ensure proven quality.
    \item Quality check by DB InfraGO AG: In the final step, the data was randomly checked by DB InfraGO AG. About 5\% of the data was reviewed.
\end{enumerate}

\begin{table}
  \centering
  \caption{Object classes annotated within the dataset}
  \begin{tabular}{|l || r | r |}
    \hline
    \textbf{Object class} & \textbf{GAF vehicle} & \textbf{BR472} \\
    \hline
    Person & 984818 & 467063 \\
    Personal Item & 70467 & 16551 \\
    Crowd & 13188 & 4044 \\
    Track & 738833&  522621 \\
    Ignore Track&	7963&	20767 \\
    Switch&	79811&	112093\\
    Train& 	98057&	102878\\
    Train Front&	26842&	32674\\
    Wagons& 	39531&	24613\\
    Buffer Stop&	15166&	25252\\
    Signal Pole&	326273&	312362\\
    Signal Bridge&	23862&	10885\\
    Signal&	148189&	447573\\
    Catenary Pole&	1494205&	516902\\
    Group of Bicycle&	5745&	1253\\
    Motorcycle&	2620&	-\\
    Road Vehicle&	175788&	127544\\
    Bicycle&	22622&	11576\\
    Pram&	4929&	4826\\
    Wheelchair&	225&	-\\
    Animal&	7473&	3971\\
    \hline
  \end{tabular}
  \label{goas}
\end{table}

\section{Multi-sensor dataset.}

\begin{figure}[t]
      \includegraphics[width=0.9\textwidth]{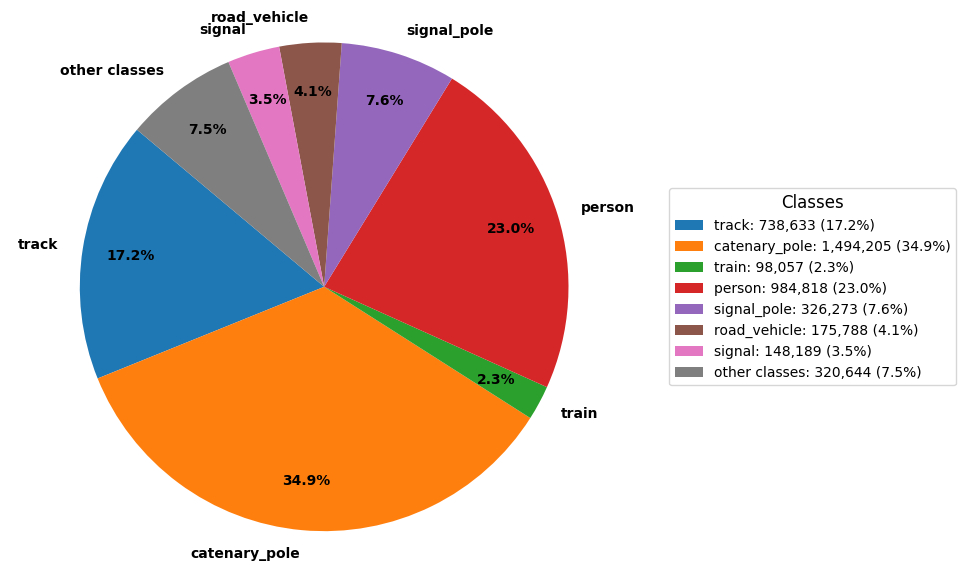}
      \caption{Disribution of annotations across classes for GAF data.}
      \label{anno-commuter-pie}
\end{figure}

\begin{figure}[t]
      \includegraphics[width=\textwidth]{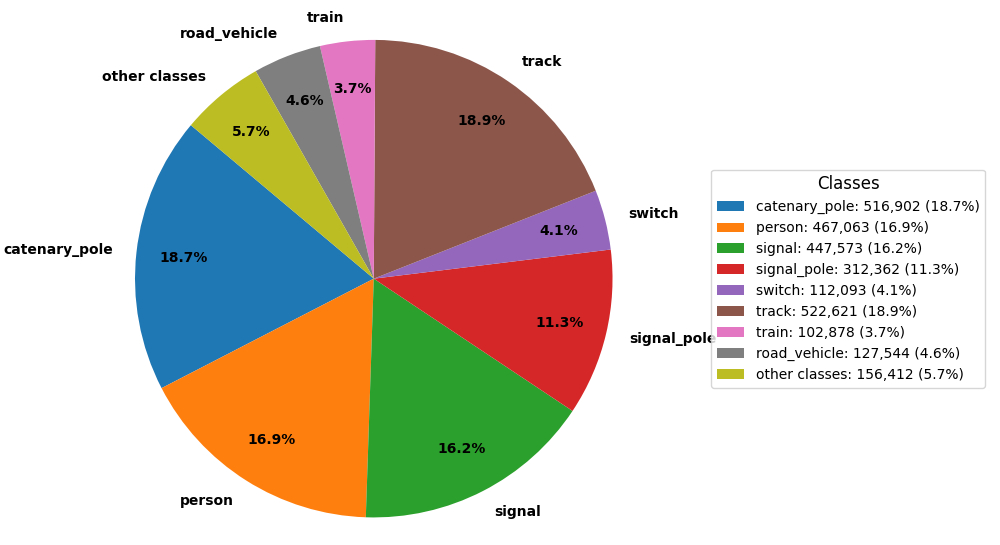}
      \caption{Disribution of annotations across classes for commuter train data.}
      \label{anno-gaf-pie}
\end{figure}

\begin{figure}[t]
      \includegraphics[width=\linewidth]{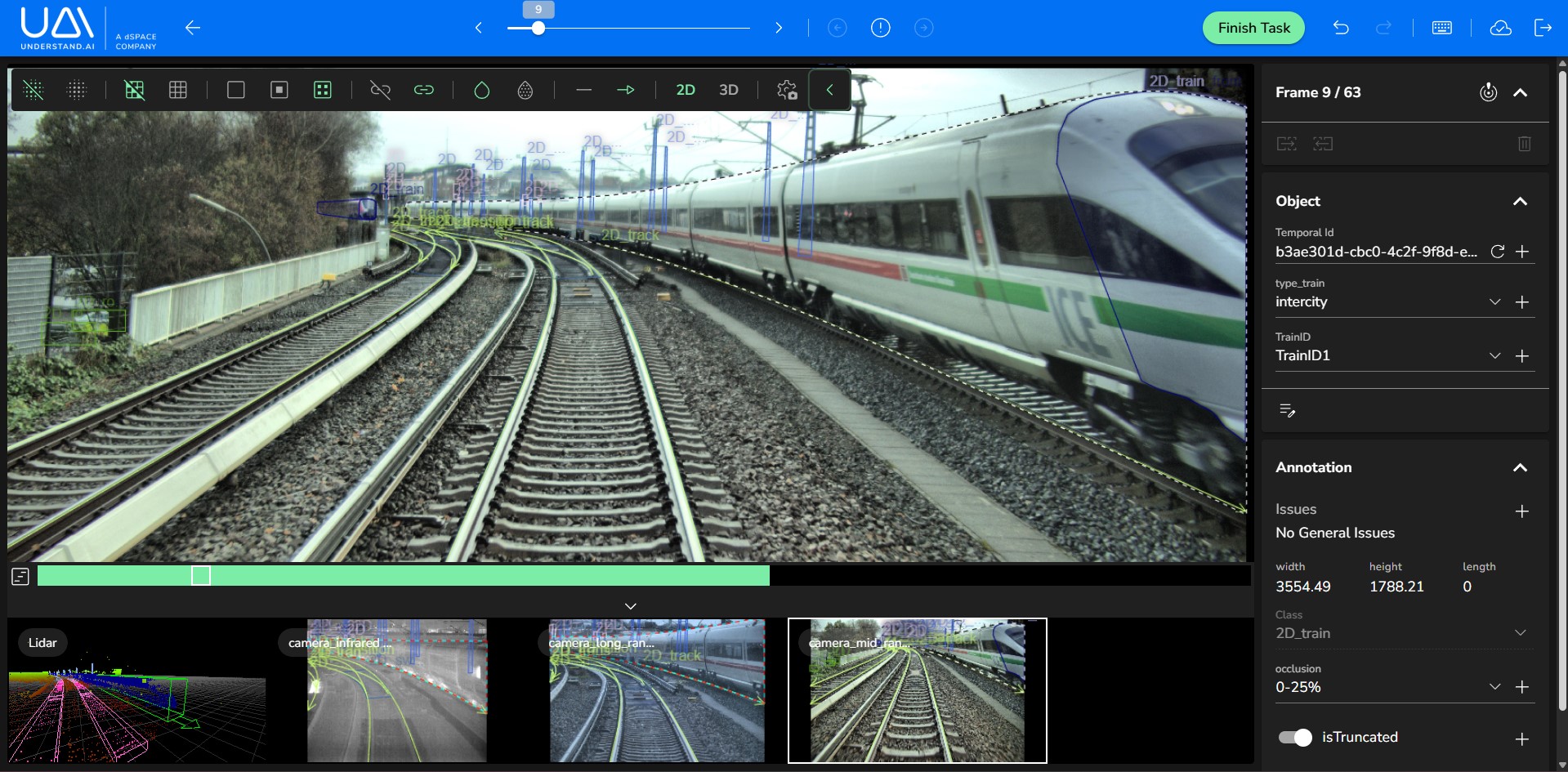}
      \caption{Train annotated with a polygon and tracks annotated with polylines in the 2D image.}
      \label{annotation-train}
\end{figure}
\begin{figure}[t]
      \includegraphics[width=\linewidth]{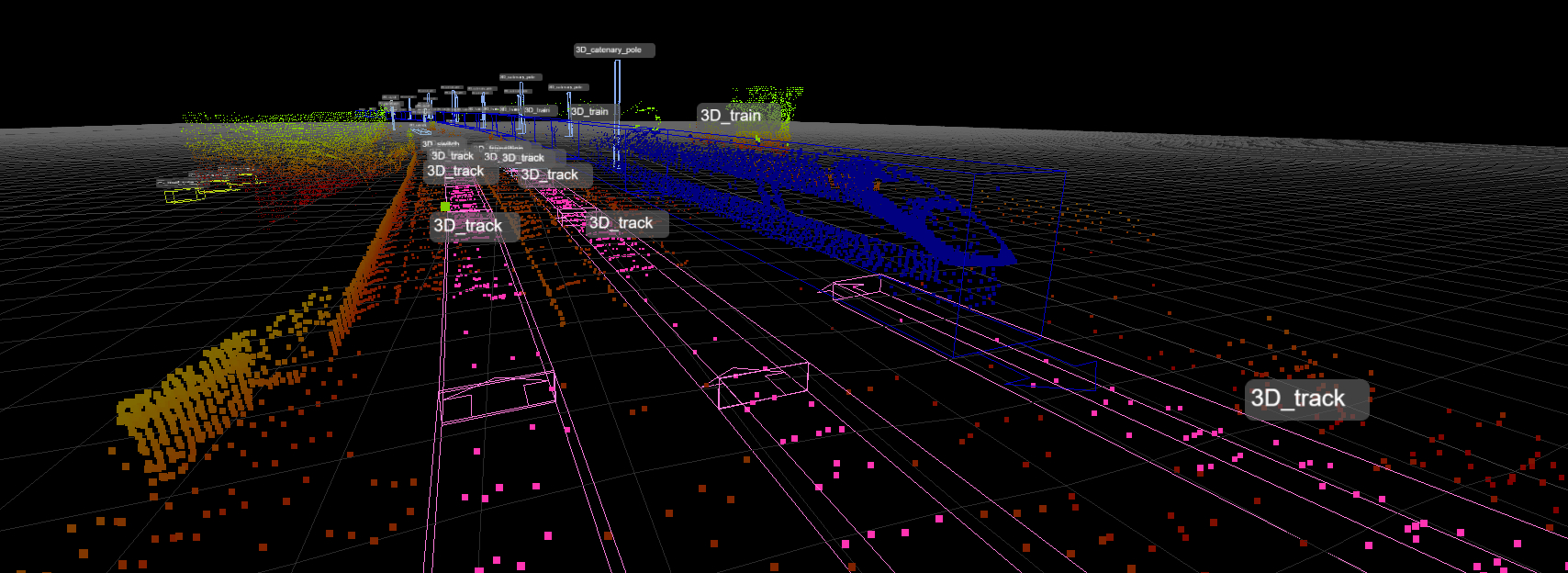}
      \caption{Train and tracks annotated with 3D bounding boxes in the 3D point clouds.}
      \label{annotation-train-3D}
\end{figure}

The created dataset comprises a total of 88.2 min (5292 s) of annotated sensor data. This includes 1981 seconds and 69 sequences of real sensor data from the GAF vehicle and 3311 seconds and 194 sequences of real sensor data from the commuter train (BR472). The dataset contains a total of 7\,052\,055
annotations. The detailed breakdown of the annotations per object class is shown in Table \ref{goas}.
Some frames of our dataset are presented in Fig. \ref{abb-1}, \ref{abb-2}, \ref{annotation-train} and \ref{annotation-train-3D}. For visualizing annotations in sensor data, the understandAI tool was used. 

The pie charts in Fig. \ref{anno-gaf-pie} and \ref{anno-commuter-pie} show the percentage distribution of annotations for the respective vehicles, i.e. GAF and commuter. Object classes that were represented less than 2\% in the annotations were summarized in “other classes”. As can be seen in the charts most of the annotations are distributed among typical railroad vehicles, railroad infrastructure elements and persons. In this way, the data set stands out from typical data sets from the automotive sector.
The annotations are provided in JSON-files that follow the RailLabel JSON schema. The RailLabel schema \cite{r15} is a subschema of the ASAM OpenLABEL standard \cite{r16}.

The dataset can be used for a wide variety of applications in areas such as GoA2 - GoA4, infrastructure monitoring and environment observation.

\section{Conclusions}

A new multi-sensor dataset for the railway sector was presented. The dataset comprises 88.2 min (5292 s) of annotated sensor data. The data was recorded as part of cooperation and pilot projects. 
The annotations were carried out by understandAI. Based on the collaboration between DB InfraGO AG and UAI, an efficient process was set up to deliver large amounts of high quality data. At the same time, the flexibility to adapt data formats and specifications to the development of future sensor configurations, project needs and requirements was maintained. It is possible to use the dataset for various applications. DB InfraGO AG has thus carried out important pioneering work and sent a clear signal to the sector. The dataset can be made available to interested parties and partners as a basis for future further development by the industry. The dataset is available upon request from DB InfraGO AG by contacting either \mbox{philipp.neumaier@deutschebahn.com} or \mbox{martin.koeppel@deutschebahn.com}.

\end{document}